\documentclass[11pt]{article}
\usepackage[a4paper]{geometry}
\usepackage{clic2016}
\usepackage{times}
\usepackage{url}
\usepackage{latexsym}
\usepackage[utf8]{inputenc}
\usepackage{todonotes}
\usepackage{scrextend}
\usepackage{caption}

\title{Italy goes to Stanford: \\
a collection of CoreNLP modules for Italian}

\author{
  Alessio Palmero Aprosio \\
  Fondazione Bruno Kessler \\
  Trento, Italy \\
  {\tt aprosio@fbk.eu} \\\And
  Giovanni Moretti \\
  Fondazione Bruno Kessler \\
  Trento, Italy \\
  {\tt moretti@fbk.eu} \\
}

\date{}

\begin{document}

\def\arraystretch{1.2}
\captionsetup{font=small}

\maketitle
\begin{abstract}
  \textbf{English.} In this we paper present Tint, an easy-to-use set of fast, accurate and extendable Natural Language Processing modules for Italian.
  It is based on Stanford CoreNLP and is freely available as a standalone software or a library that can be integrated in an existing project.
\end{abstract}

\begin{abstract-alt}
 \textrm{\bf{Italiano.}} In questo articolo presentiamo Tint, una collezione di moduli semplici, veloci e personalizzabili per l'analisi di testi in Italiano.
 Tint è basato su Stanford CoreNLP e può essere scaricato gratuitamente come software stand-alone o come libreria da integrare in progetti esistenti.
\end{abstract-alt}

\section{Introduction}

In recent years, the Natural Language Processing (NLP) technologies have become a fundamental basis for complex tasks, such as Question Answering, Event Identification and Topic Classification.
While most of the NLP tools freely available on the web (such as Stanford CoreNLP\footnote{\url{http://stanfordnlp.github.io/CoreNLP/}} and OpenNLP\footnote{\url{https://opennlp.apache.org/}}) are designed for English and sometimes adapted to other languages, there is a lack of this kind of resources for Italian.

In this paper, we present Tint, a suite of ready-to-use modules for NLP that is:

\begin{description}
\item[New.] Tint is the first completely free and open source tool for NLP in Italian.
\item[Simple.] Tint can be downloaded and used out-of-the-box (see Section~\ref{sec:tool}). In addition, it relies on Stanford CoreNLP Java interface, therefore it can be included easily into an existing project.
\item[Modular.] Tint can be extended using the CoreNLP Java interfaces. At the same time, existing modules can be replaced with more customized ones.
\item[Efficient.] In its default configuration, Tint is faster than most of its competitors (see Section~\ref{sec:eval}).
\item[Accurate.] Most of Tint modules have a state-of-the-art accuracy (see Section~\ref{sec:eval}).
\item[Free.] Tint is released as open source software under GNU GPL.
\end{description}

\section{Architecture}

The Tint pipeline is based on Stanford CoreNLP \cite{manning2014stanford}, an open-source framework written in Java, that provide most of the commons Natural Language Proccessing tasks out-of-the-box in various language.
The framework provides also an easy interface to extend the annotation to new tasks and/or languages.
Differently from some similar tools, such as UIMA \cite{Ferrucci:2004:UAA:1030318.1030325} and GATE \cite{Cunningham:2002:GAD:1073083.1073112}, CoreNLP is easy to use and does not require it to be learnt: a basic object-oriented programming skill is enough.
In Tint, we use this framework to both port the most common NLP tasks to Italian and add some new annotators for external tools, such as entity linking, temporal expression identification, keyword extraction.


\section{Modules}
\label{sec:modules}

\subsection{Tokenizer}

This module provides text segmentation in tokens and sentences.
At first, the text is grossly tokenized; in a second step, tokens that need to be put together are merged using two customizable lists of Italian non-breaking abbreviations (such as ``dott.'' or ``S.p.A.'') and regular expressions (for e-mail addresses, web URIs, numbers, dates).

\subsection{Morphological Analyzer}

The morphological analyzer module provides the full list of morphological features for each annotated token.
The current version of this module has been trained with the Morph-it lexicon \cite{Zanchetta_2005-1}, but it’s possible to extend or retrain it with other Italian datasets.
In order to grant fast performance, the model storage has been implemented with the mapDB Java library\footnote{\url{http://www.mapdb.org}} that provides an excellent variation of the Cassandra’s Sorted String Table.
To extend the coverage of the results, especially for the complex forms, such as ``porta-ce-ne'', ``portar-glie-lo'' or ``bi-direzionale'', the module tries to decompose the token into prefix-root-infix-suffix and attempts to resolve the root form.

\subsection{Part-of-speech tagger}
\label{sec:pos}

The part-of-speech annotation is provided through the Maximum Entropy implementation \cite{Toutanova:2003:FPT:1073445.1073478} included in Stanford CoreNLP.
The model is trained on the Universal Dependencies\footnote{\url{http://universaldependencies.org/}} (UD) dataset for Italian \cite{Bosco2013ConvertingIT}, a dataset -- freely available for research purpose -- containing more than 300K tokens annotated with lemma, part-of-speech and syntactic dependencies.
As an alternative, a wrapper annotator that uses TreeTagger is also available in Tint.

\subsection{Lemmatizer}

The module for the lemmatization is a rule-based system that works by combining the Part-of-Speech output and the results of the Morphological Analyzer so to disambiguate the morphological features using the grammatical annotation.
In order to increase the accuracy of the results, the module tries to detect the genre of noun lemmas relying to the analysis of their processed articles.
For instance, for the correct lemmatization of ``il latte/the milk'', the module uses the singular article ``il'' to identify the correct gender/number of the lemma ``latte'' and returns ``latte/milk'' (male, singular) instead of ``latta/metal sheet'' (female, which plural form is ``latte'').

\subsection{Named Entity Recognition and Classification}

The NER module recognize persons, locations and organizations in the text.
It uses a CRF sequence tagger \cite{Finkel:2005:INI:1219840.1219885} included in Stanford CoreNLP and it is trained on the I-CAB \cite{magnini2006cab}, a dataset  containing 180K words taken from the Italian newspaper ``L'Adige''.

\subsection{Dependency Parsing}

This module provides syntactic analysis of the text and uses a transition-based parser (included in Stanford CoreNLP) which produces typed dependency parses of natural language sentences \cite{chen2014fast}.
The parser is powered by a neural network which accepts word embedding inputs: the model is trained on the UD dataset (see Section~\ref{sec:pos}) and the word embeddings are built on the Paisà corpus \cite{lyding2014paisa}, that contains 250M tokens of freely available and distributable texts harvested from the web.

\subsection{Entity Linking}
The entity linking task consists in disambiguating a word (or a set of words) and link them to a knowledge base (KB).
The biggest (and most used) available KB is Wikipedia, and almost every linking tool relies on it.
The Tint pipeline provides a wrapper annotator that can connect to DBpedia Spotlight\footnote{\url{http://bit.ly/dbpspotlight}} \cite{isem2013daiber} and The Wiki Machine\footnote{\url{http://bit.ly/thewikimachine}} \cite{Giuliano:2009:KMM:1667988.1667993}.
Both tools are distributed as open source software and can be used by the annotator both as external services or through a local installation.

\subsection{Temporal Expression Extraction and Normalization}

The task of temporal expression extraction is included in Tint as a wrapper to HeidelTime \cite{StroetgenGertz2013:LREjournal}, a rule-based state-of-the-art temporal tagger developed at Heidelberg University. HeidelTime also normalizes the expressions according to the TIMEX3 annotation standard.
The software is released under the GPL license, therefore it can be used both for educational and commercial purposes.

\subsection{Keyword extraction}

Keyword extraction in Tint is performed by Keyphrase Digger \cite{moretti2015digging}, a rule-based system for keyphrase extraction. It combines statistical measures with linguistic information given by part-of-speech patterns to identify and extract weighted keyphrases from texts.
The CoreNLP annotator for Keyphrase Digger is included in the Tint pipeline, but the main software must be downloaded and installed from the official website\footnote{\url{http://dh.fbk.eu/technologies/kd}} as it is not released open source.

\section{Evaluation}
\label{sec:eval}

Tint includes a rich set of tools, evaluated separately.
In some cases, an evaluation based on the accuracy is not possible, because of the lack of available gold standard or because the tool outcome is not comparable to other tools' ones.

When possible, Tint is compared with existing pipelines that work with the Italian language: Tanl \cite{tanl}, TextPro \cite{pianta2008textpro} and TreeTagger \cite{Schmid94probabilisticpos}.

In calculating speed, we run each experiment 10 times and consider the average execution time.
When available, multi-thread capabilities have been disabled.
All experiments have been executed on a 2,3 GHz Intel Core i7 with 16 GB of memory.

The Tanl API is not available as a downloadable package, but it's only usable online through a REST API, therefore the speed may be influenced by the network connection.
In addition, the Tanl API does not provide offsets for the annotated text, nor it allows a text to be uploaded already tokenized and divided in sentences, therefore an automatic alignment was needed.
The tools used for this alignment are distributed as part of the Tint software.

No evaluation is performed for the Tint annotators that act as wrappers for an external tools (temporal expression tagging, entity linking, keyword extraction).

\subsection{Tokenization and sentence splitting}

For the task of tokenization and sentence splitting, Tint outperforms in speed both TextPro and Tanl (see Table~\ref{table:tokeneval}).
The number of tokens per second can be further increased by tuning the features (for example, by deactivating the regular expressions that recognize e-mail or web addresses).

\begin{table}[ht]
\centering
\small
\begin{tabular}{|l|r|}
\hline
System & Speed (tok/sec) \\
\hline
\hline
Tint & \textbf{80,000} \\
Tanl API & 30,000 \\
TextPro 2.0 & 35,000 \\
\hline
\end{tabular}
\caption{Tokenization and sentence splitting speed.}
\label{table:tokeneval}
\end{table}

\subsection{Part-of-speech tagging}

The evaluation of the part-of-speech tagging is performed against the test set included in the UD dataset, containing ~10K tokens.
As the tagset used is different for different tools, the accuracy is calculated only on five coarse-grained types: nouns (N), verbs (V), adverbs (B), adjectives (A) and other (O).
For each tool, the corresponding tagset is converted to this tagset and accuracy is calculated dividing the number of times the tagger gets the right answer by the total number of tags in the dataset.
Table~\ref{table:poseval} shows the results.

\begin{table}[ht]
\centering
\small
\begin{tabular}{|l|r|r|}
\hline
System & Speed (tok/sec) & Accuracy \\
\hline
\hline
Tint & 28,000 & \textbf{98\%} \\
Tanl API & 20,000 & n.a. \\
TextPro 2.0 & 20,000 & 96\% \\
TreeTagger & \textbf{190,000}\footnotemark & 92\% \\
\hline
\end{tabular}
\caption{Evaluation of part-of-speech tagging.}
\label{table:poseval}
\end{table}

\footnotetext{\label{ttnote}The (considerable) speed of TreeTagger includes both lemmatization and part-of-speech tagging.}

\subsection{Lemmatization}

Like part-of-speech tagging, lemmatization is evaluated, both in terms of accuracy and execution time, on the UD test set.
When the lemma is guessed starting form a morphological analysis (such as in Tint and TextPro), the speed is calculated by including both tasks.
Table~\ref{table:lemmaeval} shows the results.
All the tools reach the same accuracy of 96\% (with minor differences that are not statistically significant).

\begin{table}[ht]
\centering
\small
\begin{tabular}{|l|r|r|}
\hline
System & Speed (tok/sec) & Accuracy \\
\hline
\hline
Tint & 97,000 & 96\% \\
TextPro 2.0 & 9,000 & 96\% \\
TreeTagger & \textbf{190,000}\footref{ttnote} & 96\% \\
\hline
\end{tabular}
\caption{Evaluation of lemmatization.}
\label{table:lemmaeval}
\end{table}

\subsection{Named Entities Recognition}

For Named Entity Recognition, we evaluate and compare our system with the test set available on the I-CAB dataset.
We consider three classes: PER, ORG, LOC.
Both Tanl and TextPro deal also with the GPE class, but we merged it to LOC, as it has been done during the training of Tint.
We needed to retrain the EntityPro module of TextPro from scratch (with three classes), as the original model already contains the I-CAB test set, therefore it would overfit the results.
In training Tint, we add some gazette of names, to help the classifier to recognize entities that are not present in the training set.
In particular, we extracted a list of persons, locations and organizations by querying the Airpedia database \cite{palmero} for Wikipedia pages classified as \texttt{Person}, \texttt{Place} and \texttt{Organisation}, respectively.
The whole data used for training the NER is available for download from the Tint website.
Table~\ref{table:nereval} shows the results of the named entity recognition task.

\begin{table}[ht]
\centering
\small
\begin{tabular}{|l|r|r|r|r|}
\hline
System & Speed & P & R & F$_1$ \\
\hline
\hline
Tint & \textbf{30,000} & \textbf{84.37} & 79.97 & \textbf{82.11} \\
TextPro 2.0 & 4,000 & 81.78 & \textbf{80.78} & 81.28 \\
Tanl API & 16,000 & 72.89 & 52.50 & 61.04 \\
\hline
\end{tabular}
\caption{Evaluation of the NER.}
\label{table:nereval}
\end{table}

\subsection{Dependency parsing}

The evaluation of the dependency parser is performed against Tanl and TextPro w.r.t the usual metrics Labeled Attachment Score (LAS) and Unlabeled Attachment Score (UAS).
While Tint is trained on the UD dataset, the parsers included in Tanl \cite{attardi2013tuning} and TextPro \cite{lavelli2013ensemble} use part of the Turin University Treebank (TUT) \cite{bosco2000building}, as released for the Evalita 2011 parsing task \cite{magnini2013evaluation}.
For this reason, the comparison between the two system is not completely fair: on the one hand, the TUT dataset is smaller than the UD; on the other hand, the UD is an automatic combination of two different treebanks, that have been annotated using different guidelines \cite{Bosco2013ConvertingIT}.
Table~\ref{table:parsingeval} shows the results: the Tint evaluation has been performed on the UD test data; LAS and UAS for TextPro and Tanl is taken directly from the Evalita 2011 proceedings.

\begin{table}[ht]
\centering
\small
\begin{tabular}{|l|r|r|r|}
\hline
System & Speed & LAS & UAS \\
\hline
\hline
Tint & \textbf{9,000} & 84.67 & 87.05 \\
TextPro 2.0 & 1,300 & 87.30 & 91.47 \\
Tanl (DeSR) & 900 & \textbf{89.88} & \textbf{93.73} \\
\hline
\end{tabular}
\caption{Evaluation of the dependency parsing.}
\label{table:parsingeval}
\end{table}

\section{The tool}
\label{sec:tool}
The Tint pipeline is released as an open source software under the GNU General Public License (GPL), version 3.
It can be download from the Tint website\footnote{\url{http://tint.fbk.eu/}} as a standalone package, or it can be integrated into an existing application as a Maven dependency.

The tool is written using the Stanford CoreNLP paradigm, therefore a third part software can be integrated easily into the pipeline.
Tint accepts plain text or Newsreader Annotation Format (NAF) \cite{fokkens2014naf} as input, and CoNLL or NAF as output.


\section{Conclusion and Future Work}

In this paper we presented Tint, a simple, fast and accurate NLP pipeline for Italian, based on Stanford CoreNLP.
Currently, we offer out-of-the-box NLP annotation for part-of-speech, lemma, named entities, links to Wikipedia, dependency parsing, time expression identification and keyword extraction; additional custom modules can be added and replaced easily by implementing the CoreNLP Java interfaces.

In the future, we plan to better tune the various modules that rely on machine learning (such as dependency parsing, part-of-speech tagging and named entity recognition), that in this preliminary version of Tint have been trained without any linguistic optimization.

We are currently working on new modules, in particular Word Sense Disambiguation (WSD) w.r.t. linguistic resources such as MultiWordNet \cite{pianta2002developing} and Semantic Role Labelling, by porting to Italian resources such as Framenet \cite{baker1998berkeley}, now available in English.

On the technical side, we are updating some modules to work multi-thread.
The Tint pipeline will also be integrated into PIKES \cite{2016sac}, a tool that extracts knowledge from texts using NLP annotation and outputs it in a queryable form (such RDF triples).

\section*{Acknowledgments}

The research leading to this paper was partially supported by the European Union’s Horizon 2020 Programme via the SIMPATICO Project (H2020-EURO-6-2015, n. 692819).

\bibliographystyle{acl}
\bibliography{acl2014}

\end{document}